\documentclass[runningheads]{llncs}
\usepackage{graphicx}
\usepackage{wrapfig}
\usepackage{multirow}
\usepackage{booktabs}
\usepackage{makecell}
\usepackage{amsmath}
\usepackage{algorithm}
\usepackage{algpseudocode}
\usepackage{amssymb}
\usepackage{enumitem}
\usepackage{mathtools}
\usepackage{CJK}
\usepackage{caption}
\usepackage[normalem]{ulem}
\usepackage[T1]{fontenc}
\begin{document}
\title{Extend Adversarial Policy Against Neural Machine Translation via Unknown Token}
\titlerunning{DexChar Adversary}

\author{Wei Zou\inst{1} \and Shujian Huang\inst{1}\thanks{Corresponding Author} \and Jiajun Chen\inst{1}
}
\authorrunning{W.Zou, et al.}
\institute{National Key Laboratory for Novel Software Technology, Nanjing University, Jiangsu 210046, China\\
\email{zouw@smail.nju.edu.com} \\
\email{\{huangsj, chenjj\}@nju.edu.cn}
}
%
\maketitle              
\begin{abstract}
Generating adversarial examples contributes to mainstream neural machine translation~(NMT) robustness.  
However, popular adversarial policies are apt for fixed tokenization, hindering its efficacy for common character perturbations involving versatile tokenization.
Based on existing adversarial generation via reinforcement learning~(RL), we propose the `DexChar
policy' that introduces character perturbations for the existing mainstream adversarial policy based on token substitution.
Furthermore, we improve the self-supervised matching that provides feedback in RL to cater to the semantic constraints required during training adversaries.
Experiments show that our method is compatible with the scenario where baseline adversaries fail, and can generate high-efficiency adversarial examples for analysis and optimization of the system.

\end{abstract}

\section{Introduction}
Neural machine translation~\cite[NMT]{bahdanau2014neural,vaswani2017attention} based on the encoder-decoder framework has proved its efficacy in modern industries.
However, it's susceptible to textual noise that does not fit the training data distribution without affecting readers~(Table~\ref{tab:brittle_nmt}).
Saboteurs also craft adversarial inputs to achieve similar deterioration~\cite{belinkov2017synthetic,michel2019evaluation,zou-etal-2020-reinforced}.
Due to the black-box nature of neural networks, it is difficult to safeguard such errors, posing threats to NMT systems.
Some~\cite{karpukhin2019training,chaturvedi2019exploring,garg2020bae} handcraft adversarial tests against the NMT for system defects.
However, the extravagant expert rules can hardly cover versatile scenarios and rely heavily on expertise in the target system. 

\begin{table}[!t]
    \centering
    
    \begin{tabular}{l|l}
    \hline
    in & \begin{CJK}{UTF8}{gbsn}本工作 收到 了 许多 投诉 ， 数量 前所未有 。\end{CJK}\\ \hline
    \hline
    out  & We have received an unprecedented number of complaints.  \\ 
    \hline
    in   &  \begin{CJK}{UTF8}{gbsn}本工作 {\textbf{们}} 收到 了 \textbf{许许多多} 投诉 ， \textbf{数}  数量 前所未有 。\end{CJK} \\ 
    \hline
    out  & he said, " we have received a lot of complaints, the number of unprecedented.  \\ 
    \hline
    \end{tabular}
    
    \caption{
    The Chinese input means "This work received an unprecedented number of complaints.", which is well-translated.
    However, textual perturbations~(highlighted in bold) trigger significant errors without affecting readers.
    }
    \label{tab:brittle_nmt}
\end{table}

\begin{table*}[htb]
    \centering
    \begin{tabular}{l|l}
    \hline
    input & I love eating pineapple and kiwifruit. \\ \hline
    tokenized & I love eating pine@@ apple and kiwi@@ fruit. \\ \hline
    perturbed input  & I love eating pi\textit{\textbf{en}}apple and k\textbf{\textit{wi}}ifuit.  \\ \hline
    tokenized perturbation   & I love eating pi@@ en@@ apple and k@@ wi@@ i@@ f@@ uit. \\ \hline
    \end{tabular}
    \caption{Character-level perturbations significantly modified the tokenization, invalidating substitution-based policy. 
    `@@' indicates partition within words.
    }
    \label{tab:unk_retok}
\end{table*}
Others turned to adversarial example generation against the target NMT.
Adversarial example generation~\cite{goodfellow2014explaining} is an annotation-preserving perturbation on input that triggers system degradation, leading to off-the-shelf cases for maintenance.
The mainstream adversaries against NMT aim to search for optimal input perturbations, modeled by substitutions given corresponding tokenization.
Some policies~\cite{morris2020reevaluating,wang2022semattack,sadrizadeh2022block,cheng2019robust,cheng2020seq2sick} focus on subwords in embedding vicinity against mainstream NMT while others~\cite{ebrahimi2017hotflip,ebrahimi2018adversarial} focus on character substitutions against character-based NMT.
The adversaries refer to corresponding search loss defined upon tokenization to guide the search.
Though these methods achieve text adversary, the challenges remain as follows:

Firstly, the text adversaries are apt for the tokenization granularity of the target system, leaving out perturbations in different editorial granularity.
Mainstream methods~\cite{morris2020reevaluating,michel2019evaluation,sadrizadeh2022block,wang2022semattack} must assume static tokenization during perturbations since the similarity-based substitution with corresponding search loss is defined upon the given tokenization granularity.
Such modeling narrows the accessible adversaries since perturbations in smaller granularity risk invalidating the optimum search once it changes the overall tokenization~(Table~\ref{tab:unk_retok}).
Therefore, they can hardly reconcile perturbations across versatile granularity.
That is, adversaries against mainstream NMT modeled by tokenized subwords~\cite{sennrich2015neural} lack compatibility for common character perturbations such as homonyms~\cite{zhang2023character} and synonyms~\cite{belinkov2017synthetic,le2020detecting}. 
A practical adversarial example generation requires expanding policies for perturbations in different granularity.

Secondly, it's hard to model the semantic constraints involving perturbations across different granularity to maintain output annotations. 
Unlike minor perturbations against the image classification, minor perturbations on characters do not correspond to similarity features based on embedding of off-the-shelf subwords when they significantly modify overall tokenization~(Table~\ref{tab:unk_retok}). 
Though humans neglect their impact on semantics, mainstream deep learning modules tend to trigger \textit{false-negative} classifications given significantly modified tokenization.

This paper extends the adversary~\cite[RL-attacker]{zou-etal-2020-reinforced} based on reinforcement learning~\cite[RL]{sutton2018reinforcement}.
Specifically, RL-attacker is trained by RL, where the NMT is an environment for rewards to constrain annotation-reserving semantics and promote adversary. 
The value estimation during RL circumvents handcrafted search loss given static tokenization in the mainstream adversaries.
In this work, we propose the `DexChar' policy to introduce dexterous character perturbations to the mainstream substitution-based adversarial policy.
we additionally incorporate noisy data augmentation for the corresponding discriminator in environments, thus catering to false-negative classification, and providing compatible rewards for character perturbations during the RL.

The main contributions of this work can be summarized as follows:
\begin{itemize}
\item  We propose the `DexChar' policy which extends mainstream substitution-based adversaries with dexterous character perturbations across different granularity;
\item Meanwhile, we implement noisy data augmentation to improve semantic constraints for adversary training to circumvent \textit{false-negative} discrimination triggered by character perturbations;
\item We conduct adversaries on mainstream NMT with different input module settings. 
The results show that our improved policy is compatible with the scenarios where the mainstream adversarial policy fails, and can intuitively analyze and maintain the model without losing much efficiency.
\end{itemize}

\section{Preliminary}

\begin{figure*}[ht]
    \centering
    \includegraphics[width=1.0\textwidth]{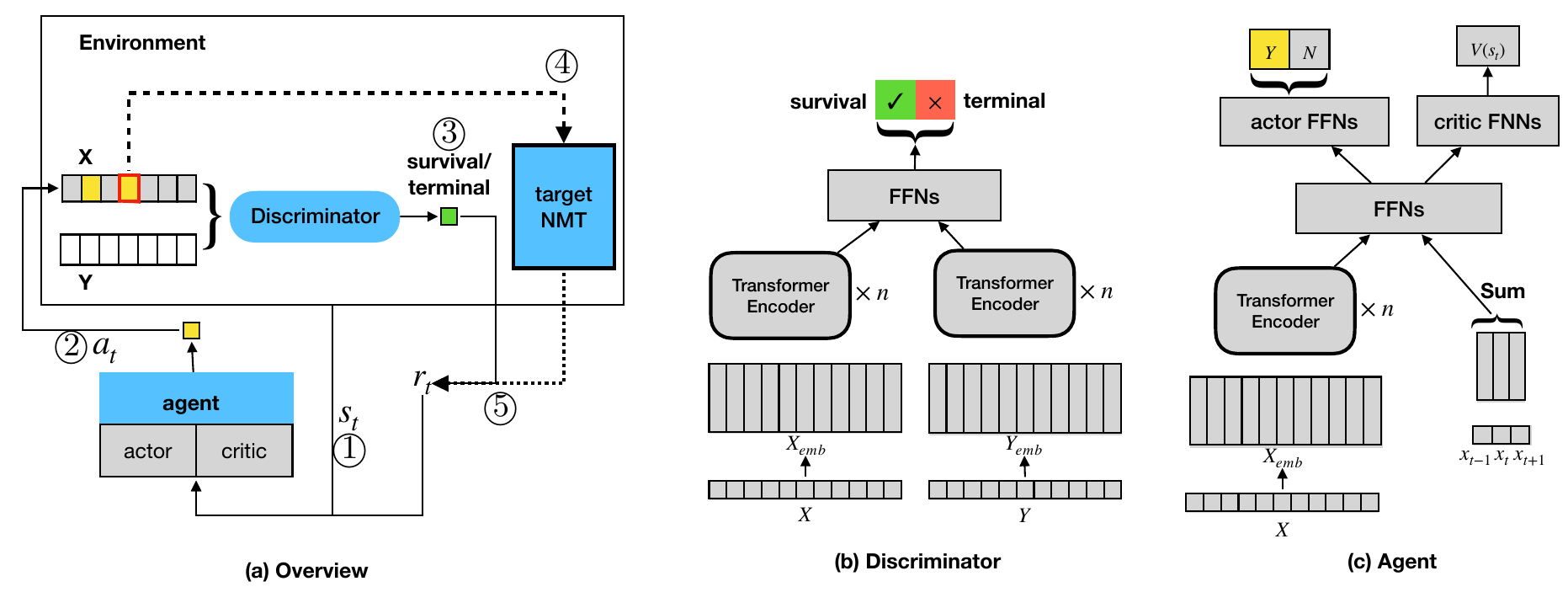}
    \caption{
    The overview of RL-attacker~\cite{zou-etal-2020-reinforced}.
    (a) The overall algorithm loops the following steps:
    \textcircled{1} agent uses the current text as state input~($s_t$) to determine whether the current position $t$~(framed in red) needs perturbation.
    \textcircled{2} The agent perturbs text if needed~(highlight in yellow) by a candidate token substitution;
    \textcircled{3} $D$ determines whether perturbation is within semantic constraint: negative~(red), the remaining perturbation is terminated with a punishment of $-1$; 
    Otherwise, the `survive' perturbation~(green) is rewarded by the positive probability of $D$, and continuing follow-up perturbations;
    Loop \textcircled{1} to \textcircled{3} until the policy survives to complete the sequential perturbations.
    \textcircled{4} The perturbed sequence is re-segmented and tested by the target NMT for episodic reward;
    \textcircled{5} Accumulate rewards $r$ and update the agent.
    (b) The architecture of discriminator $D$, with $n$ layers of Transformer encoder for input feature extraction. 
    (c) The architecture of actor-critic agent with shared input feature extraction.}
\label{fig:model_arch}
\end{figure*}

\subsection{Annotation-preserving Adversaries for Text}
Mainstream text adversaries are modeled by constrained perturbations.
Though some policies~\cite{sooksatra2024adversarial} adopt the perturbations in the latent vicinity, they will fail constraints within more delicate semantics. 
Therefore, major text adversaries~\cite{morris2020reevaluating,wang2022semattack,sadrizadeh2022block,cheng2019robust,cheng2020seq2sick} focus on token substitution in embedding vicinity.
Given the tokenization with the corresponding word embedding representation, the adversary collects all the k-nearest neighbors by cosine-similarity for each token in the vocabulary as candidates, then keeps the candidates within a vicinity radius $\epsilon$ for perturbation.
Others~\cite{ebrahimi2017hotflip,ebrahimi2018adversarial,han2022text} focus on character substitutions against character-based NMT, e.g., transposition, insertion, substitution, and deletion. 
They rely on the morphology and phonetics of characters to ensure the reader's perception of the semantics under noise~\cite{han2022text}, thus apt for maintaining annotations without embedding similarity.

Theoretically, arbitrary text perturbations can correspond to a specific substitution.
However, an applicable adversary must enumerate accessible perturbations for search loss defined upon a given tokenization as guidance.
Since it's impossible to enumerate underlying perturbations beforehand, it fails to grasp defects by various perturbation granularity.

\subsection{Adversarial Policy against NMT via Reinforcement Learning} \label{sec:rl-attack}
Adversarial example generation via RL~\cite[RL-attacker]{zou-etal-2020-reinforced} trains the adversary by interacting with the target NMT to maximize the metric degradation.
It inherits the mainstream adversarial substitution while circumventing the handcrafted search loss via value estimation for arbitrary guidance.
RL-attacker involves consecutive GAN-style~\cite{goodfellow2014explaining} updates of an \textit{environment} and an \textit{agent}~(Figure~\ref{fig:model_arch}), where the environment contains a discriminator $D$ for semantic constraints.
For $n_a$ rounds of policy updates, the environment's $D$ is updated up to $n_e$ rounds until its validation performance $\rho$ reaches $\bar\rho$.
Since the learned value estimation by critics provides policy guidance for training instead of a handcrafted search loss on a given tokenization, the corresponding policy is theoretically no longer limited to a given static tokenization and expert priors.

However, the RL-attacker only adopts the similarity-based substitution for the policy without covering character-level adversaries. 
Meanwhile, the discriminator $D$ for semantic constraints is a neural net upon direct text embedding, thus prone to yield false-negative discrimination if the character-level perturbation significantly modifies the tokenization.
$D$ tends to negate character perturbations even within semantic constraints due to tokenization issues, thus hindering underlying character adversarial policy training.

\section{Methodology}

In this section, we detail our methods with the corresponding motivations.
First, we propose the `DexChar' policy, which introduces dexterous character perturbations of different granularity by harnessing the `unknown' token~(noted as `UNK') for the existing substitution-based adversarial policy. 
Then, we implement noisy data augmentation for the corresponding self-supervised discriminator in the environment to alleviate the false-negative semantic discrimination for the extended policy.

\subsection{`DexChar' Policy} \label{sec:UNK}


The major issue in introducing arbitrary perturbations for policy is to present an entry for innumerable underlying substitutions.
Note that perturbations of different granularity impact systems by introducing \textbf{low-frequency} character combinations as shown in Table~\ref{tab:unk_retok} which disturb the language model upon trained tokenization, we propose to add candidate `UNK' to all substitutions as the entry for various underlying perturbations.
Once a token adopts the `UNK' for perturbation, generating its \textbf{valid low-frequency substitution} introduces perturbations of different granularity.
Note that 'UNK' indicates low-frequency rather than unrecognized text, since low-frequency character combinations are ubiquitous although the SOTA tokenization algorithms present arbitrary character strings.
Therefore, the DexChar policy is still viable against systems with the SOTA tokenization algorithms.

\begin{algorithm}[ht]
\caption{`DexChar' Perturbation $\Delta(w,ACT,kwargs)$}\label{alg:perturb}
\begin{algorithmic}
\Require $w=[c_0,c_1,\dots,c_n]$, target vocabulary $V$, $ACT\in\{Swap,Ins,Sub\}$
\Ensure perturbed $w_{final}$ via $ACT$.   
\For{$c$ in $w$}
\State perturb $c$ by $ACT$, generating $w'$;
\If{$w'$ is regarded `UNK' by $V$} break;
\EndIf
\EndFor
\If{$w'$ is regarded `UNK' by $V$} 
    \State success, return by $w_\text{final}=w'$
\Else
    \State failed, return by $w_\text{final}=$None
\EndIf
\end{algorithmic}
\end{algorithm}

We propose the function $\Delta(w, ACT, kwargs)$ on token $w$ defined by Algorithm~\ref{alg:perturb}, where `kwargs' introduces additional inputs such as a dictionary for substitution candidates.
Inspired by work~\cite{ebrahimi2018adversarial,zhang2023character}, we listed $ACT$ by following character perturbations:
\begin{itemize}
    \item Swap (Swap): flip the adjacent character (e.g. noise $\rightarrow$ n\textbf{io}se);
    \item Insertion (Ins): Insert a character into a word. To ensure semantics, the corresponding character is repeated (e.g. noise $\rightarrow$ no\textbf{i}ise);
    \item Substitution (Sub): Given character substitution candidate dictionary $cands_{knn}$, replaces a character within a word with another candidate character (e.g. noise $\rightarrow$ nois$\epsilon$ ).
    We follow work~\cite{ebrahimi2017hotflip} to adopt keyboard vicinity and homoglyphs as candidates.
\end{itemize}
For each input token $w$, the function $\Delta(\cdot)$ traverses each character from left to right, then perturbs the character by $ACT$ until a low-frequency token regarded as `UNK' by vocabulary $V$ is produced. 
The $\Delta(\cdot)$ yields `None' if the specified $ACT$ fails.
Note that we leave out the deletion as it's intuitively detrimental for semantics.

\begin{algorithm}[ht]
\caption{`UNK' Generation}\label{alg:unk_gen}
\begin{algorithmic}
\Require $w=[c_0,c_1,\dots,c_n]$, target vocabulary $V$, $ACT\in\{Swap,Ins,Sub\}$, dictionary of character-level substitution candidates $\text{cands}_\text{knn}$, dictionary of homophone substitution candidates $\text{cands}_\text{phone}$, perturbation function $\Delta(\cdot)$.
\Ensure perturbed $w_{final}$ which is regarded `UNK' by $V$.  
    \State $w_\text{final}=$None
    \If{homonym is available for $w$}  
        \State $w_\text{final}=\Delta(w,Sub, \text{cands}_\text{phone})$ 
    \EndIf
    \If{$w_\text{final}$ is None} 
        \If{$|w|>3$}   
            \State $w_\text{final}=\Delta(w,Swap)$ 
        \Else 
            \State $w_\text{final}=\Delta(w,Sub,\text{cands}_\text{knn})$
        \EndIf
    \EndIf
    \State $w_\text{tmp}=w$
    \While{$w_\text{final}$ is None}
        \State $w_\text{tmp}=\Delta(w_\text{tmp},Ins)$ 
        \State $w_\text{final} = w_\text{tmp} \text{if} w_\text{tmp}$ is not None
    \EndWhile
\end{algorithmic}
\end{algorithm}

To ensure the generation of a `UNK', we propose the Algorithm~\ref{alg:unk_gen}, which combines multiple $ACT$s to ensure that a valid `UNK' for the target dictionary is generated.

\subsection{Noisy Data Augmentation for Discriminator}
It is worth Noting that self-supervised $D$ intends to model the \textit{semantic matching} between the perturbed input and the original annotation rather than the \textit{variation} of texts.
We augment the positive data with a small amount of semantic-preserving character perturbation modeled by Sec~\ref{sec:UNK} to cater to false-negative discrimination triggered by perturbations of various granularity.

The augmentation adopts a dynamic ratio $\xi$, which follows the intuition to adjust with the confidence $\rho$ of on-the-fly environment $D$: 
When the confidence of the current $D$ is high, that is, $\rho>\bar\rho$, which indicates the $D$ is sufficient against perturbations, thus the training supports noisy data augmentation.
Therefore, the input $X$ of the positive sample adopts random character perturbations by the probability $\xi =\rho-\bar\rho$;
When the confidence of the environment is low, that is, $\rho <\bar\rho$, it indicates the $D$ is temporarily less capable with only standard training.
All training data for the environment's discriminator is retokenized once it's perturbed.

\section{Experiments}
We conduct all experiments on the mainstream NMT settings on $4$ NVIDIA V100~(32G) given WMT14 English-German~(en-de, 4.5M) and CWMT17 English-Chinese~(en-zh, 7.4M) parallel data.
We adopt the mainstream NMT model by end-to-end autoregressive Transformers as the backbone for adversaries.
The experiments include different embedding settings as test scenarios for verification.


\subsection{Experiment Settings}
\paragraph{Target NMT Settings} The experiments include the following embedding settings for the target NMT model which provides different robustness against input perturbations:
\begin{itemize}
    \item Base setting (base): source and target languages with independent vocabularies and embedding layers;
    \item shared-vocabulary setting~(shared-emb): source and target languages share vocabulary and embedding layers. This setting facilitates the shared semantic features between similar languages~\cite{wu2023beyond} during training~(e.g., en-de), and is adopted by mainstream multilingual applications;
    \item Pre-trained setting (PLM-emb): source and target languages adopt fixed embedding from a pre-trained language model~(PLM), whereas the remaining parameters of the NMT are trained by translation.
    This setting is popular for facilitating the semantic features extracted by mass language model training.
    In this work, we adopt the embedding of the multilingual anti-noise pre-training model mBART~\cite{liu2020multilingual}.
    Since the mBART is trained by a multilingual vocabulary, it essentially belongs to a special shared-vocabulary setting.
    \item The `unknown' vocabulary~(UNK) setting: We follow the baseline transformer training setting to truncate the top 30k vocabulary by frequency in the training dataset, with the others regarded as low-frequency for UNK.
\end{itemize}

The training settings including learning rates and model architectures follow that of the Transformer~\cite{vaswani2017attention}.

\paragraph{Training Data} The NMT training and adversarial experiments adopt the same parallel data, i.e., WMT14 en-de~(4.5M) and CWMT17 en-zh~(7.4M) data.
When training the target NMT, the en-de language pair adopts wmt13 as the validation set and wmt14 as the test set; the en-zh language pair adopts wmt17 as the validation set and wmt18 as the test set. 
The parallel data also initialize the RL environment during training adversarial generation policies.
The experiments then verify the translation performance affected by adversaries on the test set.

Text data is processed by the mosesdecoder scripts\footnote{https://github.com/moses-smt/mosesdecoder/tree/master/scripts/tokenizer}, and Chinese is further tokenized by jieba\footnote{https://github.com/fxsjy/jieba}.
Then, the NMT adopts the subword-nmt\footnote{https://github.com/rsennrich/subword-nmt} to generate subwords statistically as tokens after concatenating the source and target language data.
For the systems adopting mBART for input vocabularies and embedding, we directly adopt the tokenization and dictionary of mBART.
Due to the oversized input dictionary of mBART, we only retain the top 50k tokens by token frequencies via traversing the bilingual training data, with the rest of the vocabulary as unknown tokens, i.e., `UNK'.

\paragraph{Evaluation Metric} 
We record the average adversarial candidates~(candidate fertility, CF) of the 500 most frequent source tokens to indicate the different test scenarios for adversaries.
Intuitively, more candidates facilitate RL training for adversarial policy.
We first adopt BLEU~\cite[sacreBLEU]{post2018call}\footnote {https://github.com/mjpost/sacrebleu} for policy training, refering to the metrics in work~\cite{michel2019evaluation,zou-etal-2020-reinforced}. 
Then, we propose to adopt the following metrics to evaluate adversarial example generation:
\begin{itemize}
    \item Metric degradation~(MD): The more translation degradation after perturbation compared to the original test. 
    We turn to BLEURT~\cite{sellam2020bleurt,pu2021learning}\footnote{The BLEURT20 checkpoint, https://github.com/google-research/bleurt} for degradation, rather than the BLEU score used during training for less biased validation.
    \item Degradation per edit~(DPE): The MD is divided by the edit distance compared to the original input after the perturbation, where we adopt TER from sacreBLEU as the edit distance.
    DPE indicates the efficiency of the adversarial policy;
    \item Pairing Accuracy~(PA): The semantic matching rate between the perturbed input and the original annotation.
    The experiments adopt GPT-3.5-turbo, the application interface of the large language model~\cite[LLM]{chatgpt} by the corresponding "system" and "user" prompts shown in Table~\ref{tab:prompts} as queries.
    To validate the LLM's discrimination for noisy inputs, we randomly sample an additional 100 pairs from every experiment for human evaluation~(acc).
\end{itemize}

\begin{table}[ht]
\centering
\begin{tabular}{cc}
\hline
 & Prompt \\ \hline
Sys: & \begin{tabular}[c]{@{}l@{}}You are a knowledgable multi-lingual specialist who can tell whether  \\
           two sentences (might be different languages) are semantically matched by  \\
            "yes" or "no"\end{tabular} \\
\hline
User: & \textbf{\textless{} source\textgreater \textless{} annotation\textgreater } are they semantically matched? \\ 
\hline
\end{tabular}
\caption{Prompts for pairing accuracy}
\label{tab:prompts}
\end{table}

The ideal adversarial example generation must achieve as much metric degradation as possible within the semantic constraints that maintain the annotations, i.e., high MD with high PA. 
A low PA score indicates failed validation by mismatched annotation, instead of failed translation.
Higher DPE indicates a more efficient adversary given less perturbation against the target NMT, thus the corresponding adversarial examples are more detrimental.

\paragraph{Baseline adversaries}
\begin{itemize}
    \item Random noise injection~\cite[RNI]{ebrahimi2017hotflip,ebrahimi2018adversarial}: Randomly perturb each token given a probability $0.2$ with character perturbations defined in Sec~\ref{sec:UNK} regardless of semantics. 
    Note that RNI acts as a sanity baseline for the adversarial examples;
    \item Gradient search~\cite[GS]{michel2019evaluation}: Greedily search the substitution candidates from left to right for the perturbation combination that maximizes the adversarial loss. 
    The policy adopts vicinity candidates within the radius of the NMT's embedding space;
    \item Reinforced attack~\cite[RL-attacker]{zou-etal-2020-reinforced} in Sec~\ref{sec:rl-attack}
\end{itemize}
Experiments adopt 10 nearest candidates to calculate the vicinity radius $\epsilon$.
To sum up, RNI adopts the character perturbation for all tokens regardless of semantics;
GS only adopts substitution candidates within $\epsilon$;
RL-attacker adopts the candidates in GS and limited UNK substitution for those without candidates.
Our work supports both perturbations from GS and RNI.

\subsection{Experiment Results}

\begin{table*}[ht]
\centering
\begin{tabular}{l|lll|lll|lll}
\hline
zh -\textgreater en & \multicolumn{3}{l|}{base(MC=2.27)} & \multicolumn{3}{l|}{shared\_vocab(MC=1.06)} & \multicolumn{3}{l}{bart\_emb(MC=5.61)} \\ \hline
BLEURT              & 60.395    &         &              & 61.12       &             &                 & 59.36      &           &               \\
                    & MD        & DPE     & PA(acc)      & MD          & DPE         & PA(acc)         & MD         & DPE       & PA(acc)       \\ \hline
RNI                 & 23.955    & 1.248   & 0.62(1.0)    & 28.99       & 1.584       & 0.45(1.0)       & 29.16      & 1.793     & 0.56(0.99)    \\
GS                  & 6.945     & 0.285   & 0.85(0.99)   & 4.92        & 0.615       & 0.99(0.98)      & 9.24       & 0.462     & 0.76(1.0)     \\
RL-base             & 13.275    & 0.606   & 0.92(1.0)    & 5.82        & 0.803       & 0.91(0.98)      & 24.13      & 1.207     & 0.89(0.99)    \\
Ours                & 23.92     & 1.127   & 0.91(0.98)   & 15.92       & 1.532       & 0.92(0.99)      & 22.03      & 1.615     & 0.88(0.97)    \\ \hline
en -\textgreater zh & \multicolumn{3}{l|}{base(MC=1.12)} & \multicolumn{3}{l|}{shared\_vocab(MC=1.06)} & \multicolumn{3}{l}{bart\_emb(MC=5.61)} \\ \hline
BLEURT              & 66.79     &         &              & 67.44       &             &                 & 66.715     &           &               \\
                    & MD        & DPE     & PA(acc)      & MD          & DPE         & PA(acc)         & MD         & DPE       & PA(acc)       \\ \hline
RNI                 & 38.56     & 1.600   & 0.63(1.0)    & 31.32       & 1.559       & 0.32(0.98)      & 28.505     & 1.308     & 0.48(0.98)    \\
GS                  & 12.42     & 0.776   & 0.88(0.98)   & 13.22       & 0.925       & 0.87(1.0)       & 20.505     & 0.760     & 0.85(0.98)    \\
RL-base             & 22.67     & 1.139   & 0.74(1.0)    & 21.32       & 1.191       & 0.78(0.99)      & 22.595     & 1.224     & 0.79(1.0)     \\
Ours                & 40.98     & 1.722   & 0.8(0.98)    & 27.2        & 1.386       & 0.82(0.97)      & 27.395     & 1.179     & 0.82(1.0)     \\ \hline
\end{tabular}
    \caption{Experiment results for zh-en language pair. $\uparrow$ indicates `the more the better'. `CF' indicates the candidate's fertility for the substitution adversary.
    The `acc' indicates the sampled human validation for GPT-3.5-turbo's paring accuracy~(PA). 
    Note that RNI with an excessively lower semantic pairing~(PA) is not a valid adversarial example generation.
    }
    \label{tab:main_experiment_zhen}
\end{table*}

The results of en-zh and en-de language pairs are shown in Tables~\ref{tab:main_experiment_zhen} and~\ref{tab:main_experiment_ende}, respectively.
Note that the high accuracy by human evaluation~(acc) verifies the LLM's evaluation~(PA).
The results are summarized as follows:
\begin{itemize}
    \item
    Fewer candidates~(CF) indicate more difficulty against adversaries by substitution thus less degradation~(MD). 
    For instance, GS achieves significantly less MD for both en-zh and en-de NMT with shared vocabulary.
    RNI and our work with character perturbation circumvent these scenarios with more MD and more efficient adversary~(DPE).
    Note that RNI with significantly lower PA scores indicates its inability to follow semantics constraints even with high MD, which are not always valid adversarial examples.
    \item
    Given the same substitution candidates for adversaries, our policy achieves better semantic constraints to maintain annotation, i.e., MD with a higher PA score.
    RNI is regarded as the sanity baseline whose adversary injects the random noise regardless of semantics for the original annotation.
    Therefore, RNI easily achieves significant degradation by failed validation with mismatched annotation, whereas our method reconciles the adversary~(MD) within semantics constraints~(PA).
    
    \item 
    Compared with the original RL-attacker, our method achieves a better adversary~(MD, DPE) due to the extensive DexChar policy, which introduces character perturbations without failing semantic constraints~(PA). 
    DexChar reaches a similar adversary compared to the sanity baseline RNI with much more reasonable semantics~(PA) required by adversarial examples. 
\end{itemize}

Notably, mBART embedding does not significantly hinder the adversary, whereas its embedding structure facilitates the mainstream substitution adversary with more candidates~(CF).  

\begin{table*}[htb]
\centering
\begin{tabular}{l|lll|lll|lll}
\hline
en -\textgreater de & \multicolumn{3}{l|}{base(MC=3.34)} & \multicolumn{3}{l|}{shared\_vocab(MC=1.01)} & \multicolumn{3}{l}{bart\_emb(MC=7.48)} \\ \hline
BLEURT              & 70.44     &         &              & 71.02        &            &                 & 71.025     &           &               \\
                    & MD$\uparrow$        & DPE$\uparrow$     & PA(acc)$\uparrow$      & MD$\uparrow$           & DPE$\uparrow$        & PA(acc)$\uparrow$         & MD$\uparrow$         & DPE$\uparrow$       & PA(acc)$\uparrow$       \\ \hline
RNI                 & 42.46     & 3.45   & 0.42(1.0)    & 49.52        & 3.668      & 0.49(1.0)       & 39.62     & 2.386     & 0.58(1.0)     \\
GS                  & 29.3      & 2.402   & 0.83(0.98)   & 11.62        & 2.152      & 0.95(0.98)      & 9.805      & 0.649     & 0.92(1.0)     \\
RL-attacker             & 38.34     & 2.506   & 0.89(1.0)    & 29.92        & 2.267      & 0.92(0.99)      & 20.015     & 1.409     & 0.90(1.0)     \\
Ours                & 48.67     & 3.380   & 0.91(0.97)   &51.905      & 3.767      & 0.91(0.99)      & 30.21      & 2.464     & 0.92(1.0)     \\ \hline
de -\textgreater en & \multicolumn{3}{l|}{base(MC=4.35)} & \multicolumn{3}{l|}{shared\_vocab(MC=1.01)} & \multicolumn{3}{l}{bart\_emb(MC=7.48)} \\ \hline
BLEURT              & 70.86     &         &              & 71.61        &            &                 & 71.15      &           &               \\
                    & MD$\uparrow$        & DPE$\uparrow$     & PA(acc)$\uparrow$      & MD$\uparrow$           & DPE$\uparrow$        & PA(acc)$\uparrow$         & MD$\uparrow$         & DPE$\uparrow$       & PA(acc)$\uparrow$       \\ \hline
RNI                 & 40.74     & 3.08   & 0.48(0.98)   & 35.11        & 2.66      & 0.55(0.99)      & 38.75      & 2.266     & 0.59(1.0)     \\
GS                  & 7.135     & 0.469   & 0.88(0.97)   & 10.49        & 2.185      & 0.96(0.99)      & 7.95       & 0.520     & 0.87(0.97)    \\
RL-attacker             & 29.24     & 2.215   & 0.9(0.98)    & 23.27        & 1.907      & 0.92(0.99)      & 25.93      & 1.826     & 0.93(0.96)    \\
Ours                & 42.475    & 2.909   & 0.89(1.0)    & 36.28       & 2.232      & 0.93(0.97)      & 39.10     & 2.330     & 0.92(0.99)    \\ \hline
\end{tabular}
    \caption{Experiment results for en-de language pair}
    \label{tab:main_experiment_ende}
\end{table*}
\section{Analysis}
In this section, we take the Chinese-to-English translation with shared vocabulary as an example, since the Chinese character poses a larger impact on semantics, to analyze adversarial example generation, including adversarial generation efficiency, translation defect analysis, and follow-up system maintenance.

\begin{figure}[htb]
    \centering
    \includegraphics[width=0.5\textwidth]{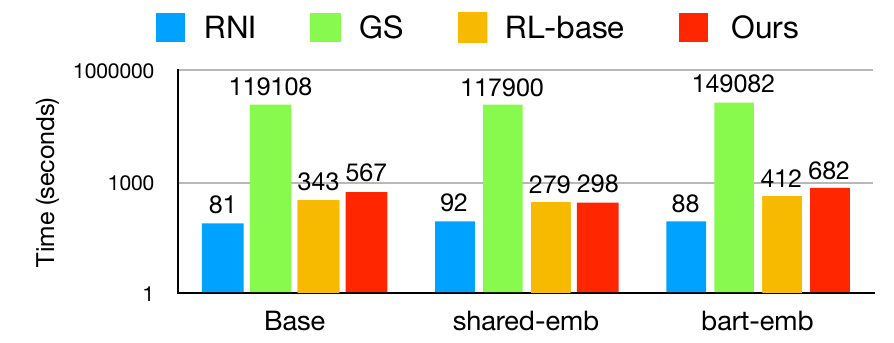}
    \caption{Overheads for different adversaries. Our method retains the adversarial efficiency with minor overhead for extended character-level perturbations.}
    \label{fig:efficiency}
\end{figure}

\subsection {Adversarial Generation Efficiency}
Figure~\ref{fig:efficiency} records the overhead~(seconds) for adversarial generation.
GS calculates the gradient features as search guidance for candidates, thus resulting in a large overhead.
Our method retains the efficiency of the RL paradigm with over $200$ times acceleration compared to GS, while the extended character-level perturbations for policy take up only negligible overheads.

\begin{figure*}[ht]
    \centering
    \includegraphics[width=0.9\textwidth]{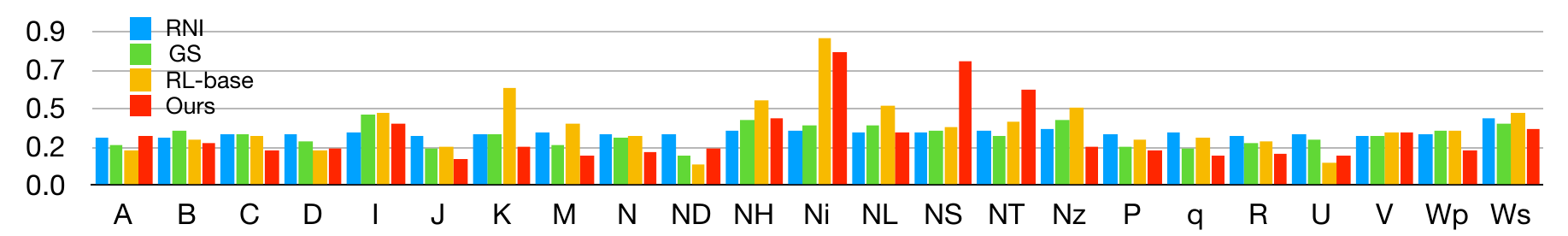}
    \caption{Preference over POS by different adversaries}
    \label{fig:pattern}
\end{figure*}

\subsection{Adversarial Pattern}
Handcrafted adversaries~\cite{garg2020bae,samanta2017towards} prefer to perturb tokens with more potential impacts on semantics.
Intuitively, the policy trained by RL~\cite{zou-etal-2020-reinforced} also prefers perturbation for specific sentence components~(POS).
The analysis collects the perturbation ratio of the adversaries against each type of POS tag~(POS tags by 863 standard~\footnote{https://catalog.ldc.upenn.edu/docs/LDC2012T05/readme.html}).
As shown in Figure~\ref{fig:pattern}, our method retains the preference among certain POS by RL-attacker~\cite{zou-etal-2020-reinforced} with a slight difference.
Due to the extended character perturbations, our method achieves more adversaries against POS such as temporal noun~(NT) and geographical name~(NS) as their semantics are less sensitive to character perturbations such as homonyms. 
On the other hand, our methods extracted fewer adversaries against POS such as conjunctions~(C), proverbs~(I), numbers~(M), and quantifiers~(q), whose semantics are largely posed at the character level. 
The preference indicates that our extended policy can identify and utilize character perturbations for more effective and versatile adversaries.

\subsection {Fine-tuning with Adversarial Examples}
We fine-tune the target model with the corresponding adversarial examples.
Due to the efficiency of the RL adversary, we adopt the parallel data $<X,Y>$ to generate the same amount of adversarial pairs $< X', Y>$ for fine-tuning by the following loss function base:
\begin{equation}
     \Bar {L} = {L}(X, Y) +\lambda {L}(X ', Y),
\end{equation}
where $\lambda$ is the adversarial coefficient, which we set to $0.2$, and $L$ is the end-to-end supervised loss of NMT.
The fine-tuning follows the training settings by NMT~\cite{vaswani2017attention} for up to one epoch with a fixed learning rate at $1e-6$.
We test the robustness of NMT against the following noisy scenarios:
\begin{itemize}
    \item RNI~(0.1): Random noise injection with a low probability at $0.1$. Perturbations are defined in Sec~\ref{sec:UNK};
    \item RL-attacker: Adversarial tests generated by the RL-attacker;
    \item Mixed: Adversarial tests generated by character-level perturbation and token substitution;
\end{itemize}
The results are shown in Table~\ref{tab:tuning}.
Intuitively, each fine-tuning improves the corresponding test scenario.
However, fine-tuning is potentially detrimental to other scenarios.
Notably, the RL-attacker does not improve the robustness against character perturbations, whereas our method can mildly alleviate the noise by character perturbation as well as adversarial inputs. 

\begin{table}[htb]
    \centering
    
\begin{tabular}{lllll}
\hline
          & None    & RNI (0.1) & RL-attacker & Mixed \\ 
\hline
None        & 23.1 & 16.4      & 19      & 16.5 \\
RL-attacker FT  & 22.7(-0.4) & 16.1(-0.3)      & 22.6(+3.6)    & 17.3(+0.8) \\
DexChar FT    & 21.9(-1.2) & 18.5(+2.1)      & 21.3(+2.3)    & 22.1(+5.6)\\
\hline
\end{tabular}

\caption{Adversarial fine-tuning for zh$\rightarrow$en NMT with shared vocabulary.
The table columns present test scenarios, while the rows present the corresponding fine-tuned~(FT) setting.
The "None" represents the original NMT model and the original test scenario.}
\label{tab:tuning}
\end{table}
\section{Conclusion}
This work proposes the `DexChar' policy, which utilizes the `UNK' token for dexterous substitution, extending the adversarial policy to accommodate versatile perturbation granularity. 
Experiments show that the perturbations introduced by the `DexChar' policy facilitate the adversary against test scenarios where the traditional adversary fails.
The resulting adversary remains efficient and available against versatile target models, enabling subsequent analysis.
Adversarial tuning with our method can cater to noise maintenance across different perturbation granularity.

\section*{Acknowledgement}
This work is supported by the National Science Foundation of China (No. 62376116, 62176120), and the Fundamental Research Funds for the Central Universities (No. 2024300507).


%
%
%
\bibliographystyle{splncs04}
\bibliography{ccmt_sim}

\begin{thebibliography}{10}
\providecommand{\url}[1]{\texttt{#1}}
\providecommand{\urlprefix}{URL }
\providecommand{\doi}[1]{https://doi.org/#1}

\bibitem{bahdanau2014neural}
Bahdanau, D., Cho, K., Bengio, Y.: Neural machine translation by jointly learning to align and translate. arXiv preprint arXiv:1409.0473  (2014)

\bibitem{belinkov2017synthetic}
Belinkov, Y., Bisk, Y.: Synthetic and natural noise both break neural machine translation. arXiv preprint arXiv:1711.02173  (2017)

\bibitem{chaturvedi2019exploring}
Chaturvedi, A., KP, A., Garain, U.: Exploring the robustness of nmt systems to nonsensical inputs. arXiv preprint arXiv:1908.01165  (2019)

\bibitem{cheng2020seq2sick}
Cheng, M., Yi, J., Chen, P.Y., Zhang, H., Hsieh, C.J.: Seq2sick: Evaluating the robustness of sequence-to-sequence models with adversarial examples. In: Proc. of AAAI (2020)

\bibitem{cheng2019robust}
Cheng, Y., Jiang, L., Macherey, W.: Robust neural machine translation with doubly adversarial inputs. arXiv preprint arXiv:1906.02443  (2019)

\bibitem{ebrahimi2018adversarial}
Ebrahimi, J., Lowd, D., Dou, D.: On adversarial examples for character-level neural machine translation. arXiv preprint arXiv:1806.09030  (2018)

\bibitem{ebrahimi2017hotflip}
Ebrahimi, J., Rao, A., Lowd, D., Dou, D.: Hotflip: White-box adversarial examples for text classification. arXiv preprint arXiv:1712.06751  (2017)

\bibitem{garg2020bae}
Garg, S., Ramakrishnan, G.: Bae: Bert-based adversarial examples for text classification. arXiv preprint arXiv:2004.01970  (2020)

\bibitem{goodfellow2014explaining}
Goodfellow, I.J., Shlens, J., Szegedy, C.: Explaining and harnessing adversarial examples. arXiv preprint arXiv:1412.6572  (2014)

\bibitem{han2022text}
Han, X., Zhang, Y., Wang, W., Wang, B., et~al.: Text adversarial attacks and defenses: Issues, taxonomy, and perspectives. Security and Communication Networks  (2022)

\bibitem{karpukhin2019training}
Karpukhin, V., Levy, O., Eisenstein, J., Ghazvininejad, M.: Training on synthetic noise improves robustness to natural noise in machine translation. In: Proceedings of the 5th Workshop on Noisy User-generated Text (W-NUT 2019) (2019)

\bibitem{le2020detecting}
Le, T., Park, N., Lee, D.: Detecting universal trigger's adversarial attack with honeypot. arXiv preprint arXiv:2011.10492  (2020)

\bibitem{liu2020multilingual}
Liu, Y., Gu, J., Goyal, N., Li, X., Edunov, S., Ghazvininejad, M., Lewis, M., Zettlemoyer, L.: Multilingual denoising pre-training for neural machine translation. TACL  (2020)

\bibitem{michel2019evaluation}
Michel, P., Li, X., Neubig, G., Pino, J.M.: On evaluation of adversarial perturbations for sequence-to-sequence models. arXiv preprint arXiv:1903.06620  (2019)

\bibitem{morris2020reevaluating}
Morris, J.X., Lifland, E., Lanchantin, J., Ji, Y., Qi, Y.: Reevaluating adversarial examples in natural language. arXiv preprint arXiv:2004.14174  (2020)

\bibitem{chatgpt}
OpenAI: https://openai.com/blog/chatgpt (2022)

\bibitem{post2018call}
Post, M.: A call for clarity in reporting bleu scores. arXiv preprint arXiv:1804.08771  (2018)

\bibitem{pu2021learning}
Pu, A., Chung, H.W., Parikh, A.P., Gehrmann, S., Sellam, T.: Learning compact metrics for mt. In: Proceedings of EMNLP (2021)

\bibitem{sadrizadeh2022block}
Sadrizadeh, S., Dolamic, L., Frossard, P.: Block-sparse adversarial attack to fool transformer-based text classifiers. In: Proc. of ICASSP (2022)

\bibitem{samanta2017towards}
Samanta, S., Mehta, S.: Towards crafting text adversarial samples. arXiv preprint arXiv:1707.02812  (2017)

\bibitem{sellam2020bleurt}
Sellam, T., Das, D., Parikh, A.P.: Bleurt: Learning robust metrics for text generation. In: Proceedings of ACL (2020)

\bibitem{sennrich2015neural}
Sennrich, R., Haddow, B., Birch, A.: Neural machine translation of rare words with subword units. arXiv  (2015)

\bibitem{sooksatra2024adversarial}
Sooksatra, K., Khanal, B., Rivas, P.: On adversarial examples for text classification by perturbing latent representations. arXiv preprint arXiv:2405.03789  (2024)

\bibitem{sutton2018reinforcement}
Sutton, R.S., Barto, A.G.: Reinforcement learning: An introduction. MIT press (2018)

\bibitem{vaswani2017attention}
Vaswani, A., Shazeer, N., Parmar, N., Uszkoreit, J., Jones, L., Gomez, A.N., Kaiser, {\L}., Polosukhin, I.: Attention is all you need. Proc. of NeurIPS  (2017)

\bibitem{wang2022semattack}
Wang, B., Xu, C., Liu, X., Cheng, Y., Li, B.: Semattack: Natural textual attacks via different semantic spaces. arXiv preprint arXiv:2205.01287  (2022)

\bibitem{wu2023beyond}
Wu, D., Monz, C.: Beyond shared vocabulary: Increasing representational word similarities across languages for multilingual machine translation. arXiv preprint arXiv:2305.14189  (2023)

\bibitem{zhang2023character}
Zhang, S., Wu, H., Zhu, G., Xin, X., Su, M.: Character-level adversarial samples generation approach for chinese text classification. Journal of Electronics and Information Technology  (2023)

\bibitem{zou-etal-2020-reinforced}
Zou, W., Huang, S., Xie, J., Dai, X., Chen, J.: A reinforced generation of adversarial examples for neural machine translation. In: Proc. of ACL (2020). \doi{10.18653/v1/2020.acl-main.319}

\end{thebibliography}

\end{document}